# Fuzzy Logic Model for Predicting the Heat Index

Nnamdi Uzoukwu[1, a], Acep Purqon[2, b]

[1] *Graduate of Computational Science,*
*The Faculty of Mathematics and Natural Science, Bandung Institute of Technology.*
*Jl. Ganesha No. 10, Bandung, Indonesia, 40132*

[2] *Physics of the Earth and Complex Systems Laboratory*
*The Faculty of Mathematics and Natural Science, Bandung Institute of Technology.*
*Jl. Ganesha No. 10, Bandung, Indonesia, 40132*

a) nnamdi44@students.itb.ac.id
b) acep.purqon@itb.ac.id

**Abstract**

A fuzzy inference system was developed for predicting the heat index from temperature and relative humidity data. The effectiveness of fuzzy logic in using imprecise mapping of input to output to encode interconnectedness of system variables was exploited to uncover a linguistic model of how the temperature and humidity conditions impact the heat index in a growth room. The developed model achieved an $R^2$ of 0.974 and a RMSE of 0.084 when evaluated on a test set, and the results were statistically significant ($F_{1,5915} = 222900.858$, $p < 0.001$). By providing the advantage of linguistic summarization of data trends as well as high prediction accuracy, the fuzzy logic model proved to be an effective machine learning method for heat control problems.

# 1. Introduction

The heat index is defined as the temperature that would produce equivalent stress on the body if the water vapor pressure were changed to a predetermined reference pressure [1]. It provides information about the impact of humidity on the degree of hotness and coldness of a body, and thus is the most common quantity used by environmental health experts to measure heat exposure [2]. Heat monitoring and control are important for preventing heat stress, which can cause damage to plants and humans alike.

The purpose of this work is to use fuzzy logic in the prediction of the heat index from temperature and relative humidity with a focus on heat monitoring in a green house. Chand et al. [1] applied several machine learning methods to predict the heat index from temperature and relative humidity, including artificial neural network (ANN), auto regression integrated moving average model (ARIMA) and multiple linear regression (MLR). According to the authors, ANN achieved the best performance. The efficient parameter tuning of the ANN approach offers greater resilience and thus greater accuracy. The fuzzy logic approach used in this work gave a better performance than the MLR but was not compared with high performance models like the ANN. For predictive modeling, fuzzy logic has the advantage of using linguistic variables to model system behavior so that the underlying mechanism is easy to understand and apply to control problems. The linguistic modeling of fuzzy logic provides a reasoning tool in the form of if-then rules by which the properties of a system can be understood without expert experience and rigorous analysis. By allowing imprecision in set assignment, fuzzy logic provides a way of modeling complex interaction that is adaptable.

The data used for this study was collected from a growth room where green choy sums (Brassica rapa var. parachinensis) were planted. Choy sum is a flowering herbaceous plant that is a member of the Brassicaceae, or cabbage family [4]. The outcome of this research will be useful in building expert systems for heat control in a greenhouse. The objective of this work was to develop a linguistic model of the relationship between temperature, humidity, and heat index in the growth room and apply the model to forecast the heat index, and evaluate its performance.

## 2. Methodology

### 2.1 Data Collection

The data used in developing this model was collected from a growth room in Bandung, Indonesia. The temperature and relative humidity measurements were taken at 50-second intervals over a 30-day period using a DHT11 sensor. The initial heat index readings are derived with the Adafruit library [2] based on the equations of Rothfutsz and Steadman [3]. The data attributes include relative humidity, temperature and heat index as shown in Table 1

Table 1. The data attributes used for model development.

| Measurement | Unit | Total Sample |
|---|---|---|
| R. Humidity | Percent | 23566 |
| Temperature | Celsius | 23566 |
| Heat Index | Celsius | 23566 |

### 2.2 Data Pre-Processing

The data was cleaned before it was deployed for analysis and model development. The preprocessing involved correcting inconsistent data entries and removing outliers. The interquartile range (IQR) method was employed in removing outliers to reduce noisy measurements that might have resulted from faulty sensor readings.

### 2.3 Data Analysis

Exploratory data analysis was carried out to understand the characteristics of the data set. First, the descriptive statistics was explored. Second, boxplots and histograms were used to visualize the spread and skew of the data as well as the position of outliers. Third, the predictive power of the input variables was analyzed with scatter plot and Pierson's correlation.

Table 2. Descriptive statistics of the data attributes

|  | R. Humidity | Temperature | Heat Index |
|---|---|---|---|
| Mean | 76.49 | 24.62 | 25.11 |
| Standard deviation | 2.75 | 0.54 | 0.52 |
| Min | 68.00 | 23.00 | 23.34 |
| Max | 84.00 | 26.00 | 25.70 |

Table 2 shows the range of the measurements and the average values for the data attributes. The standard deviation was highest in the relative humidity measurements and least in the temperature. The variation pattern of the observations was explored with the time series plot in figure 1.

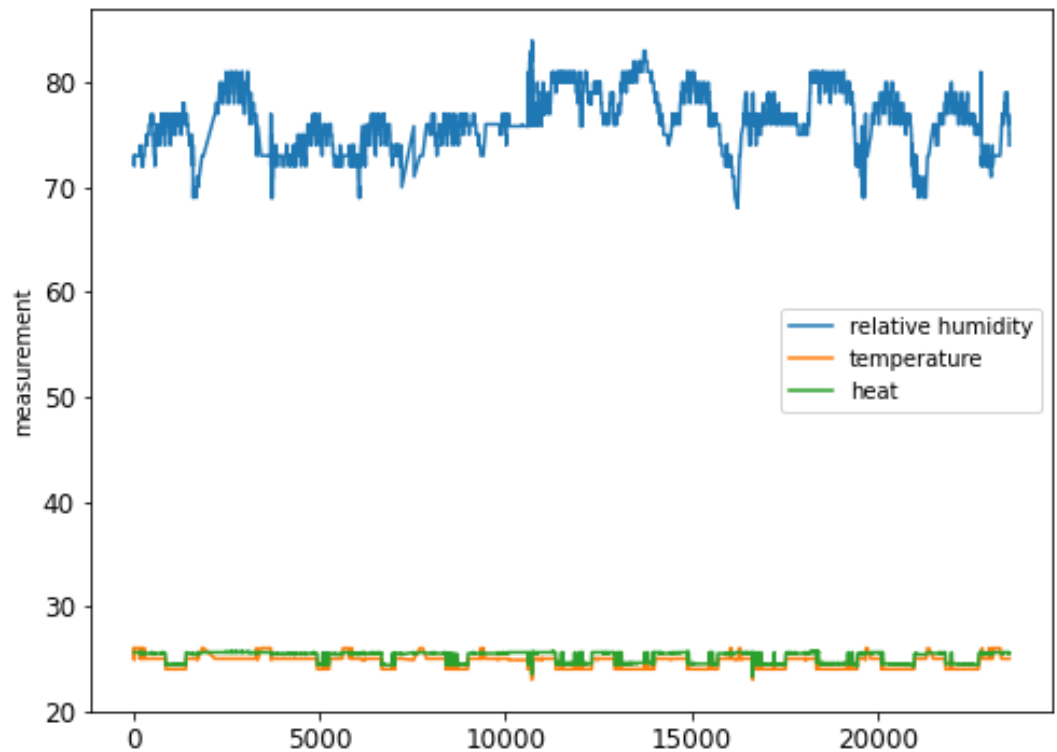

Figure 1. Time series plot of the observations for the 30-day period of measurement

The temperature and the heat index measurements showed similar variation patterns. The relative humidity measurements, however, showed a greater fluctuation.

Table 3. Pearson's correlation among the data attributes

|  | R. Humidity | Temperature | Heat Index |
| --- | --- | --- | --- |
| R. Humidity | 1.00 | -0.47 | -0.34 |
| Temperature | -0.47 | 1.00 | 0.95 |
| Heat Index | -0.34 | 0.95 | 1.00 |

The Pierson's correlation scores among the data attributes are shown in table 3. The relative humidity correlates negatively and poorly with both the temperature and the heat index, with a Pierson's correlation score of -0.47 and -0.34 respectively. The temperature correlates positively and strongly with the heat index, with a correlation score of 0.95. This demonstrates that the temperature is a good predictor of the heat index.

## 2.4　Sampling Method

The clean data, consisting of 19717 observations, was deployed for model development. It was split randomly into a train/test ratio of 70:30. The training set was used to develop the fuzzy rules and the performance of the developed model was validated on the test set.

## 2.5　Model Development

In general, a fuzzy inference system (FIS) consists of four components. The fuzzifier transforms the system's numerical inputs into linguistic values as determined by the membership functions. The knowledge base holds the data and the rules. The inference engine makes inferences about the input values according to the fuzzy rule set.

Finally, the de-fuzzifier transforms the fuzzy subsets obtained from the inference engine into numerical values [4].

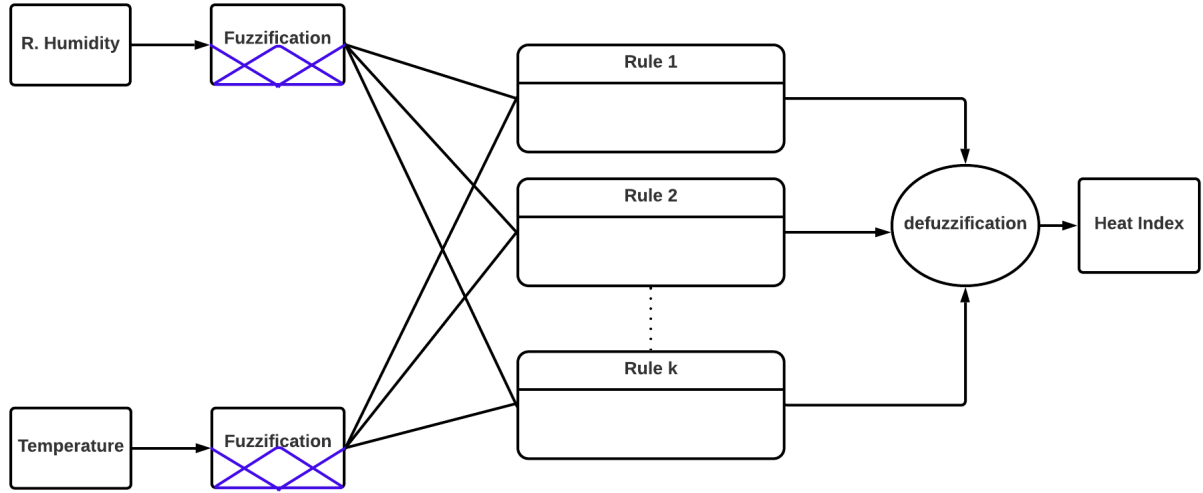

Figure 2. The structure of the proposed fuzzy inference system

Figure 2 shows the structure of the proposed fuzzy inference engine. The model will accept numerical temperature and relative humidity measurements and fuzzify them using the triangular membership functions. The fuzzy outputs are evaluated against each of the rules in the knowledge base. The inferences from the rules are accumulated and a zero-order Takagi and Sugeno approach [5] is applied to compute a crisp prediction for the heat index.

### 2.5.1 Generating Fuzzy Rules from the Data Set

The approach proposed in [6] was applied to extract fuzzy rules from the training data. Given *n* input observations of relative humidity *RH* and temperature *T*, a mapping for the heat index *HI*,

$f : (RH, T) \rightarrow HI$, was determined as follows:

- First, the universe of discourse was determined by the domain interval for the input and output variables:

$$\text{Universe} \rightarrow [\min(RH), \max(RH)], [\min(T), \max(T)] \text{ and } [\min(HI), \max(HI)]$$

- Second, the spaces of the input and output variables were divided into fuzzy regions, j, where $j = 1, 2, 3$, and represents *low, mid, and high*. Assignments to the fuzzy sets were implemented using triangular membership functions as shown in figure 3. The membership functions have the triangle formula:

$$f(x; a, b, c) = \max\left(\min\left(\frac{x-a}{b-a}, \frac{c-x}{c-b}\right), 0\right) \qquad (1)$$

- where $x$ is an element in the specified fuzzy region. The definition of $a$, $b$ and $c$ for each fuzzy region given a variable $X$ (*RH*, *T*, or *HI*) is as follows.

*low*: this was formed with a right triangle.

$a = \min(X)$

$b = \min(X)$

$c = \text{median}(\min(X), \max(X))$

*mid*: this was formed with an isosceles triangle.

$a = \min(X)$, the left vertex.

$b = \text{median}(\min(X), \max(X))$, the center.

$c = \max(X)$, the right vertex.

*high*: this was formed with a right triangle.

$a = \text{median}(\min(X), \max(X))$

$b = \max(X)$

$c = \max(X)$

- Third, the membership grades, $\mu_{A_j}(x_i)$, of the $x_i th$ observation to the fuzzy region, $A_j$, were determined. Where $i = 1, 2, \ldots, n$. Consequently, the training observations are mapped to values between 0 and 1 as follows: $\mu_{A_j}(RH_i) \to [0,1]$, $\mu_{A_j}(T_i) \to [0,1]$, and $\mu_{A_j}(HI_i) \to [0,1]$.

- Fourth, each observation was assigned to the region where it had a maximum degree of membership. For example, the first row of data ($i = 1$), could be transformed as given below.

$$(RH_1, T_1; HI_1) \Rightarrow [RH_1(A_1, max), T_1(A_2, max); HI_1(A_3, max)]$$

And it will lead to the following rule:

$$rule_1: if\ RH_1\ is\ A_1\ and\ T_1\ is\ A_2\ then\ HI_1\ is\ A_3$$

- Fifth, the degree of each rule is computed as the product of the membership grades for the input and output observations. In this regard, the degree of $rule_1$ above is given by:

$$Degree\ (rule_1) \Rightarrow \mu_{A_1}(RH_1)\mu_{A_2}(T_1)\mu_{A_3}(HI_1)$$

- Sixth, a summary of the rules was obtained by aggregating rules that have the same antecedents and selecting only one amongst them where the rule degree is highest.

$$final \text{ rules} \Rightarrow \max(rule_k)$$

Where k represents rules that has the same antecedents

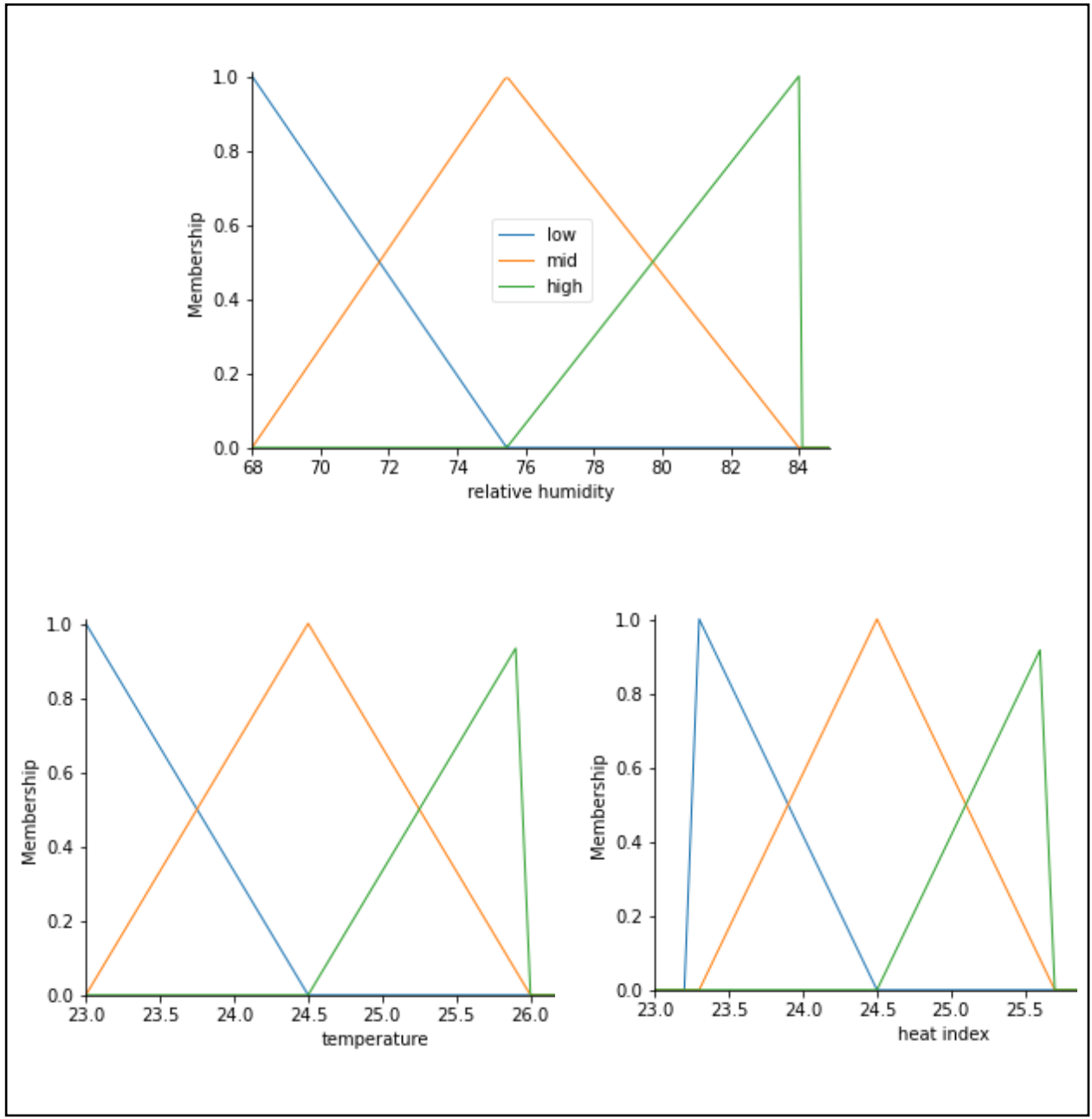

Figure 3. The Fuzzy membership functions used for model development for all the data attributes.

## 2.5.2 Defuzzification Method

Defuzzification is the process of taking fuzzified input data and turning it into a numerical output. Using the fuzzy rules to forecast the heat index from unforeseen temperature and relative humidity readings, as described in [6], involves the following steps.

- First, the input data, $(RH_i, T_i)$, were fuzzified to determine $(\mu_A(RH_i), \mu_A(T_i))$, their grade of belonging to the fuzzy region $A$.
- Second, the antecedent of the *r-th* rule was combined using a product operation to determine the degree of fulfillment [7] for each rule, $D^r$, for the *i-th* observation as follows

$$D^r = \mu_A^r(RH_i)\mu_A^r(T_i) \qquad (2)$$

- Third, the $HI_i$ was computed using the following centroid defuzzification method:

$$HI_i = \frac{\sum_{r=1}^{k} D^r \overline{HI}^r}{\sum_{r=1}^{k} D^r} \qquad (3)$$

Where $\overline{HI}^r$ was arbitrarily chosen as the center value of the fuzzy region that make up the consequent of the *r*-th rule, and $k$ is the total number of rules in the fuzzy rule set.

*Calculation of $\overline{HI}^r$ and Model Optimization*

The $\overline{HI}$ was defined as the point that has the smallest absolute value among all the points where the membership function for the fuzzy region had a membership grade equal to one [6]. Because a triangular membership function was employed, there was exactly one point where the membership grades are one at each of the fuzzy regions, which includes the following:

for region *low*; $\overline{HI}$ = min ($HI$),

for region *mid*, $\overline{HI}$ = median (min ($HI$), max ($HI$))

for region *high*, $\overline{HI}$ = max ($HI$)

The optimization strategy involved varying the $\overline{HI}^r$ within the range [-1, 1] while observing the $R^2$ and RMSE.

## 3. Discussion and Discussion

### 3.1 Control Rules

Table 4. The final fuzzy rule set

| R. Humidity | Temperature | Heat Index | Degree |
|---|---|---|---|
| high | low | low | 0.816537 |
| high | mid | mid | 0.586294 |
| low | mid | mid | 0.671715 |
| mid | high | high | 0.534828 |
| mid | low | low | 0.936416 |
| mid | mid | mid | 0.643347 |

Table 5. The fuzzy rule propositions

| Rule | Proposition |
|---|---|
| 1 | **IF** relative humidity is high and temperature is low  **THEN** heat index is low |
| 2 | **IF** relative humidity is high and temperature is mid  **THEN** heat index is mid |
|  | . . . |
| 6 | **IF** relative humidity is mid and temperature is mid  **THEN** heat index is mid |

Table 4 gives the final fuzzy rule set and the degree of significance for each rule, while Table 5 describes the application of the rules for control situations. The final fuzzy rule base comprises of six rules which summarize the strongest relationships among the data attributes. The high correlation between the temperature and the heat index, as shown by the Pierson correlation scores (table 3), is also captured in the rule set as both attributes have the same linguistic labels in the rule propositions. The rule degree shows the significance of each proposition. Rules of a higher degree will have more impact on predictive modelling and vice versa. The rule proposition that states that a mid-level humidity condition combined with a low-level temperature condition will result in a low heat index in the growth room has the highest rule degree, approximately 0.94. This rule signifies a close-to-proper assignment of the fuzzy sets while rules with degree around 0.5 signifies imprecise set assignments and helped to explain a complex relationship among the variables.

## 3.2  *Performance Evaluation*

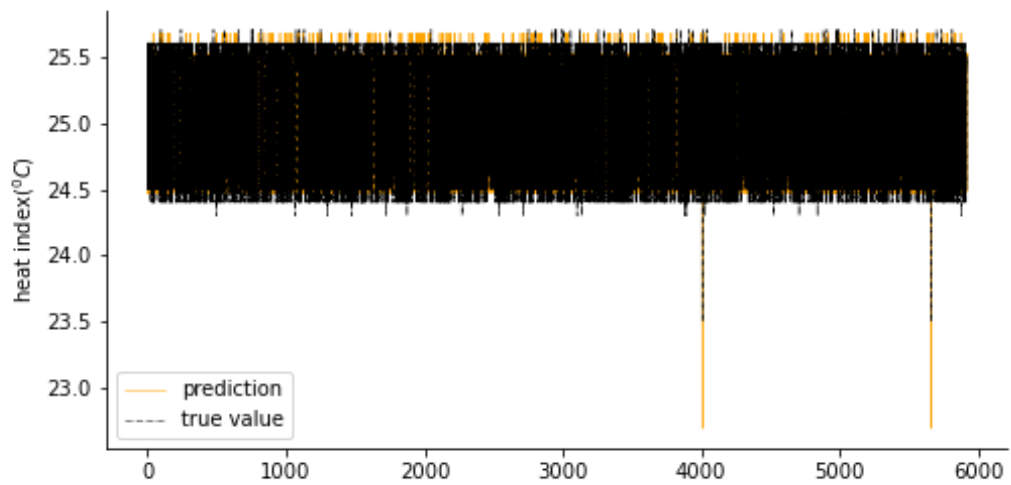

Figure 4. Prediction of the heat index for the test data *with* the fuzzy logic model

Table 6. Performance result achieved by the fuzzy logic model

| Quantity | Value | | |
|---|---|---|---|
| $R^2$ | 0.974 | | |
| RMSE | 0.084 | | |
| MAE | 0.064 | | |
| F-statistic | score | p value | df |
| | 222900.858 | 0.000 | 5915 |

According to [8], model validity should be evaluated by determining if model output agrees with observed data and whether the theory and assumptions underlying the model are justifiable. As the science behind the modeling of heat index with temperature and relative humidity is generally accepted [3], the evaluation of this model was focused on its performance on a test set comprising of 5916 observations.

Figure 5 shows the agreement between the test observations and the predicted outcome. The fuzzy logic model obviously captured the trend in the true measurements. The performance metrics for this prediction are shown in Table 6. The model was tested on the $R^2$ and RMSE while the F-statists were used to test its validity. The model achieved an $R^2$ of 0.974. $R^2$ statistics measures the proportion of variability in the true values that can be explained by the model. Scores that are close to 1, indicates that a large proportion of the variability in the response has been explained by the developed model [9]. Therefore, 97% of the variations in the test observations was explained by the fuzzy logic model. In a previous study, an $R^2$ score of 0.67 was regarded as substantial for partial least square path models [10].

The root-mean-square-error, or RMSE, of the model's prediction is 0.084. According to Gareth et al [9] the MSE is a good measure of the quality of fit—it gives an idea of the extent to which the predicted response is close to the true observation. The MSE will be small if the predicted responses are very close to the true values and large otherwise. Thus, the model shows a predictions accuracy that is less than ±0.1°C. Further, the F-test, performed at 0.05 confidence interval, proves there was a significant effect of the predicted outcome on the true measurement ($F_{1,5915}$ = 222900.858, $p < .001$)

*3.3 Error Analysis*

.

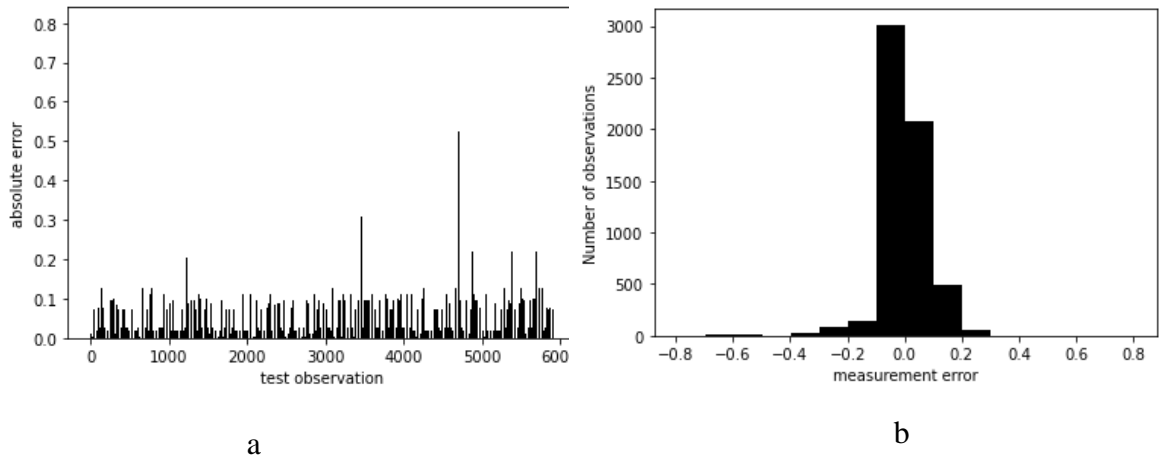

Figure 5. Distribution of the prediction errors for the fuzzy logic model (a) absolute errors at each data point (b) a histogram of the prediction errors

The model errors are largely below 0.1 °C, as indicated by the mean absolute error, MAE, of 0.064. However, a few data points have significant prediction errors (figure 5a). The observed absolute error ranges from [0.003, 0.8]. The left-skewed histogram indicates that more of the residuals have a negative value, which shows overprediction in about 56% of the observations.

## Conclusion

A fuzzy logic model constructed from triangular membership functions has been employed to extract fuzzy rules from a data set consisting of temperature, relative humidity, and heat index. The fuzzy rule set was applied in a control function to predict the heat index from unseen temperature and relative humidity data. The performance metrics of the model on the new data set are $R^2 = 0.974$, RMSE = 0.084, and MAE = 0.064. These results show a statistical significance of $F_{1,5915} = 222900.858$, $p < .001$ for the f-statistic at the 0.05 confidence interval.

The Fuzzy logic model clearly outperforms the multiple linear regressor, which achieved an $R^2$ of 0.9175 and a RMSE of 0.15 on the same test data. This demonstrates that the fuzzy set assignment was an effective method for capturing the complex non-linear dependence of the heat index on the temperature and humidity conditions. The level of prediction accuracy achieved further proves that fuzzy logic can be employed to solve heat control problems for practical purposes.